\documentclass{article}

\usepackage{arxiv}

\usepackage[utf8]{inputenc} % allow utf-8 input
\usepackage[T1]{fontenc}    % use 8-bit T1 fonts
\usepackage{hyperref}       % hyperlinks
\usepackage{url}            % simple URL typesetting
\usepackage{booktabs}       % professional-quality tables
\usepackage{amsfonts}       % blackboard math symbols
\usepackage{nicefrac}       % compact symbols for 1/2, etc.
\usepackage{microtype}      % microtypography
\usepackage{lipsum}
\usepackage{graphicx}
\usepackage{subfigure}

\title{A Privacy-preserving Distributed Training Framework for Cooperative Multi-agent Deep Reinforcement Learning}

\author{
 Yimin Shi \\
  School of Data Science\\
  The Chinese University of Hong Kong, Shenzhen\\
  Shenzhen, China\\
  \texttt{yiminshi@link.cuhk.edu.cn} \\
}

\begin{document}
\maketitle

\begin{abstract}
Deep Reinforcement Learning (DRL) sometimes needs a large amount of data to converge in the training procedure and in some cases each action of the agent may produce regret. This barrier naturally motivates different data set or environment owners to cooperate to share their knowledge and train their agents more efficiently. However, it raises the privacy concern if we directly merge the raw data from different owners. To solve this problem, we proposed a new Deep Neural Network (DNN) architecture with both global NN and local NN, and a distributed training framework. We allow the global weights to be updated by all the collaborator agents while the local weights are only updated by the agent they belongs to. In this way we hope the global weighs can share the common knowledge among these collaborators while the local NN can keep the specialized properties and ensure the agent to be compatible with its specific environment. Experiments show that framework can efficiently help agents in same or similar environments to collaborate in their training process and gain higher convergence rate and better performance.
\end{abstract}

% keywords can be removed

\section{Introduction}
Deep Reinforcement Learning (DRL) is based on the Deep Neural Network (DNN). It hopes to use the DNN model to directly estimate the value of a state through or approach the agent’s optimal policy, which can map the state to the optimal action. However, DNN models’ training generally requires a large amount of training data, which significantly increases training difficulty and may make it impossible for some small data set owners to successfully or effectively train their value function or policy model. What's more, training DRL models in some scenarios like recommendation systems will cause extra regrets if it fails to attract the user to click. Therefore, naturally and intuitively, we assume that multiple data set owners in the same or similar industry environment will have the motivation to cooperate with each other in the training procedure. Among their “similar but not necessarily the same” environments, each cooperator would hope that his own model can enjoy more data and knowledge from other cooperators' environments, and eventually achieve a model with good enough performance but with less training iterations. But at the same time, they also do not want their data privacy to be leaked. For example, let's we assume that the participants of the training are multiple companies come from a similar modern internet industry, holding their customers’ user profile data and the access to their users' clicking responses of the recommended contents. When they need to generate a recommendation system based on reinforcement learning according to their specific businesses environments, they may hope to borrow each other’s user data and user responses to complete the training process. In which their own model can be better fitted in less iterations, also with higher robustness. However, at the same time, according to internal business considerations or existing external limitations brought by the law like GPDA, they don’t want these privacy-sensitive user data to be directly accessed by other companies, especially their competitors. That is, they want to protect their own data privacy when sharing the knowledge. In this case, to solve this problem, we proposed a new distributed training framework for the cooperative multi-agent DRL which preserves the privacy of all participants.
\par
The proposed framework borrowed some ideas from the related work of Federated Learning (FL) where different training participants can be jointly trained without directly sharing the raw training data sets. Instead, participants share the model parameters and weights through a well controlled protocol. In this way, they can exchange the knowledge through the gradient but protect the privacy by keeping it at local. In our scenario, we further hope that different agents can share some similarities on the common knowledge while retaining the particularity of their specific model and environments. Because in our hypothesis, these agents involved in training must not only consider the similarity of their respective environments but also consider the difference. For example, when cooperating on the training of recommendation systems in modern internet industry, the business logic of different internet companies often share certain similarities, but since they are usually in charge of different subdivisions, they are actually located in different environments: they have different business models and user properties. To achieve this objective, we hope to train a set of neural network models for each agent, part of which are shared and updated together by every collaborator (we call it the global NN). While the other part is more private (we call it a local network), which can only be updated by the agent it belongs to and aims to learn for the private properties and ensure that the input and output of the model are compatible with the specific environment. What's more, a distributed training framework should also be introduced to transport the weights, gradients and ensure the consistency of different agents.
\par
The remaining sections of the paper are organized as follows. We review some related works in Section 2, and discuss the structure of our training framework and the NN pattern in Section 3. Section 4 discusses the privacy issue. In Section 5, we present the results of a set of experiments that explore the efficiency of the system in different agent-environment collaboration conditions. Finally, Section 6 concludes the paper and discusses the aspects that can be further studied on the basis of this work.

\section{Related Work}
\paragraph{Multi-agent Deep Reinforcement Learning}
Some article mention that learning in multi-agent environments is inherently more complicated than in single-agent situations\cite{hernandez2018multiagent}, because the agents interact with the environment and potentially each other at the same time. But in our design, there is no direct communication between agents, and all of them exchange information by and only by modifying the global NN together. Recent works on multiple-agent deep learning also have addressed the scalability and have focused significantly less on convergence guarantees.

\paragraph{Federated Learning}
Federated Learning (FL) is a distributed machine learning framework used to solve privacy issues and backbone network pressure issues\cite{yang2019federated}. Its main logic is to perform an aggregation of the model parameters of different local trainers in each round, and release it to each training participant in the next round to ensure the consistency of the new model. Some recent works\cite{shi2020edge} proposed new FL framework which can utilize the edge computing to improve the efficiency. We will apply the ideas from the privacy protection mechanism of FL on the special problem of reinforcement learning by sharing parameters instead of sharing data sets. In addition, there are some works\cite{zhuo2019federated} that focus on federated reinforcement learning, but these jobs are essentially different from ours as follows: (1) Their multi-agents in these jobs rely on each other in action, while ours do not; (2) They require different agencies to share a same environment, but we allow a certain degree of difference; (3) The specific structures of the neural network is completely different.

\paragraph{Privacy-preserving Reinforcement Learning}
In the past, there were some works\cite{sakuma2008privacy} focus on privacy-preserving reinforcement learning, but these works have two main differences from us: (1) These work are not based on deep learning, that is to say, they are not in the model considered. Deep Neural Networks (DNN); (2) At the same time, they rely on some cryptographic-based methods to protect the privacy, which is very different from the method and the federated architecture we proposed.

\section{System Design}
\subsection{Distributed Training Framework}
  In order to solve the privacy issue, we mainly hope to borrow some ideas and practices in federated learning and federated reinforcement learning, which transfers knowledge through exchanging model parameters and weights instead of directly sharing the data set. Therefore, in order to achieve the objective, we hope that each agent can share a part of its policy NN model or value function approximation NN model, which is the same to all the other agents, and make all agents jointly update this part of the NN with gradients and maintain the data consistency of weights. This part of the NN model is called global NN and is represented by the yellow part in Fig.\ref{fig1}. Its purpose is to allow different data set owners and agents to share their common knowledge. To implement this in a system, we can use a distributed training framework which where each agent can encrypt the newly generated gradient, pass it to an anonymous forwarder named Black Board (BB), which will continue to pass it to all the other agents. And for other agents, they can freely decrypt the encrypted gradient.
  
\subsection{Local and Global Neural Structure}
\begin{figure}
    \centering
    \centerline{\includegraphics[scale=0.3]{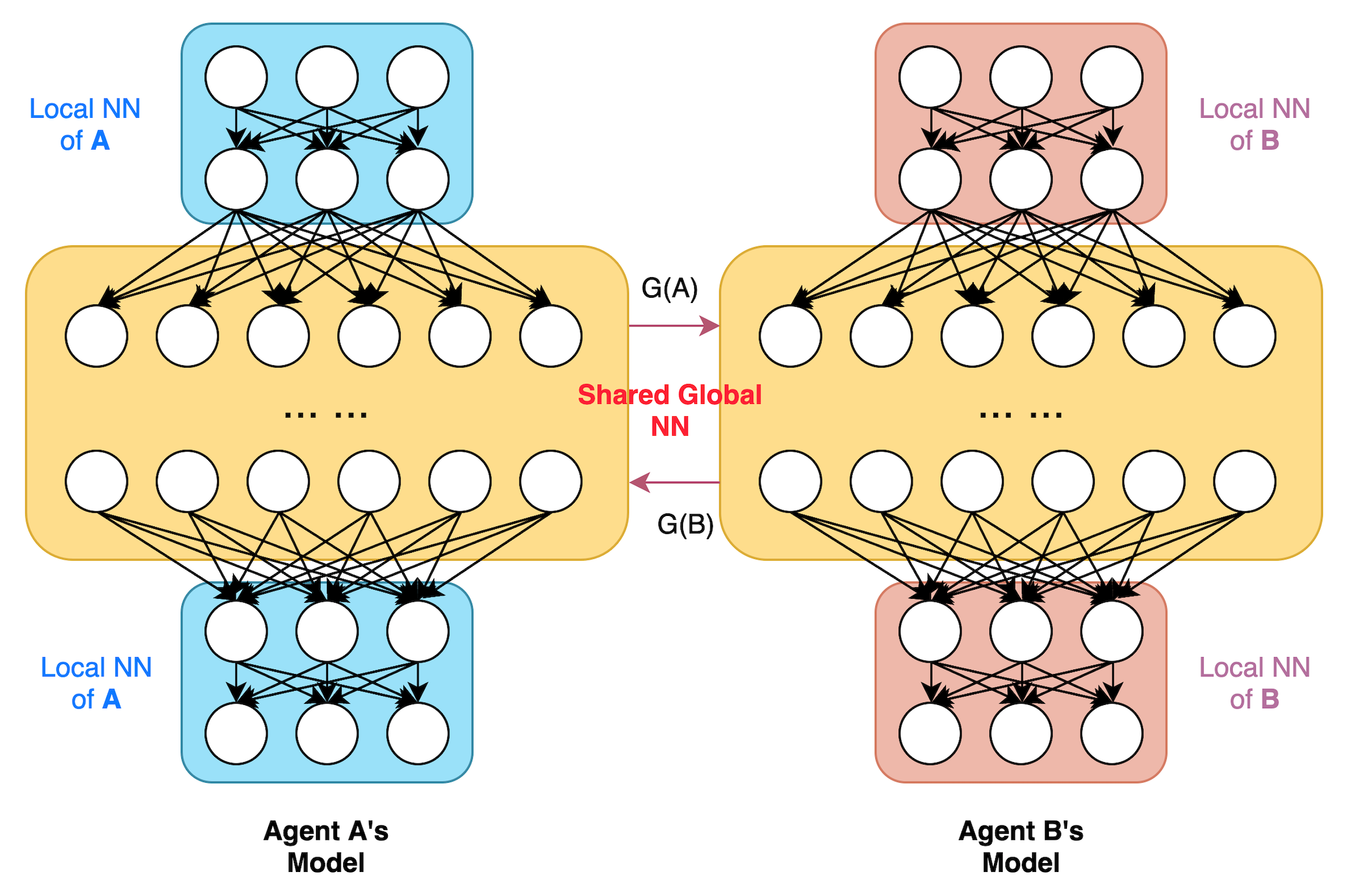}}
    \caption{Deep RL model design of multiple agents with local and global NN structure.}
    \label{fig1}
\end{figure}

  What's more, we do not assume that every agent's environment shares the same state space and action space, which means that we allow different agents to accept different input forms and decide on different , even with different dimensions. Similar to the second objective, we do not limit that the action set of each agent is the same, we allow different agents to have different actions in their respective spaces. In order to these achieve goals, we plan to design a separate local part of NN for each agent to ensure that they can be compatible with their own state space and action space, as shown in the blue and red parts of Fig.\ref{fig1}.

\section{Privacy Protection}
Whether this model can effectively protect privacy will be the most challenging part of our work, because just like research in cryptography field, it could be difficult to prove that there will be no further attacking algorithms to be create in the future, which may cause the inefficiency of the model and lead to the privacy leakage. But generally, we can at least refer to the discussion of privacy in FL to confirm our idea: first, the gradient generated by the model in the training process will not directly disclose the user's input data. Secondly, as for some typical attack models that may exist in our model, we can refer to the Deep Leakage of Gradient \cite{yang2019federated} in FL: This is a kind of attack that takes the parameters of the model as constants and the input of the model as variables. Through continuous optimization process, gradients generated by the model will be the same as the real gradient. In this way, the optimized model input will approach the real input which is originally hidden from the agent. For this issue, to which we can refer, FL already has some protection measures, such as differential privacy and model pruning. Similarly, we can use these methods to ensure the security of privacy of our own model.

% \begin{figure}
%     \centering
%     \centerline{\includegraphics[scale=0.3]{f1-shaved.png}}
%     \caption{Deep RL model design of multiple agents with local and global NN structure.}
%     \label{fig1}
% \end{figure}

\section{Experiments and Results}
\subsection{Experiments Setting}
In order to verify the effectiveness of the proposed framework, we use it to implement a simple DQN model and test it in different Gym environments. Our main test environment is CartPole. The dimension of its continuous observation space is 4, and the discrete action space's dimension is 2. In order to enable multiple agents to cooperate with each other in same or different environments, we designed the following neural network structure for 2 different agents (agent A and agent B): the network structure of agent A consists of a fully connected layer ($L_{A,1}$) with the size of (4,32), a ReLu activation layer, a fully connected layer ($L_2$) with the size of (32,16), a second ReLu activation layer, another full junction layer ($L_{A,3}$) with size of (16,2) and a final Sigmoid activation layer; Agent B has the same network structure as agent A, including three full connection layers ($L_{B,1}$, $L_2$ and $L_{B,3}$ and the same activation layers. You may noticed that we made a sequencial index for each layer (for example, $L_{A,1}$ and $L_{2}$). 

The sequence number of each local layer contains two foot-marks, which respectively indicate the agent to which the layer belongs and the position of the layer in the neural network of that agent. The only global layer $L_2$ has only one foot-mark '2', which is used to mark the it's same position in the structure of both agents.

In each epoch, we first allow agent A to interact with his environment until a failure (the definition of the 'failure' here refers to the condition that cart cannot continue to support pole in CartPole environment, for example, the inclination angle of pole is greater than a certain value). Then we allow agent B to start interacting with the environment until it fails. When both sides fail, we will conduct offline training or reset the environment according to the situation.

Due to the randomness in the training process of this kind of agent in reinforcement learning, it is difficult to judge the training quality by the change of each epoch in a single training. Therefore, in order to solve this problem, all the reward curves in this part summarize the actual reward curves of 10 times of training, that is, the average of 10 independent training results. We found that this can effectively solve the above problems, and we can clearly observe the situation of the reward curve. More than that, we added a polynomial regression for each curve so that we can better compare them in a single plotted figure.

What's more, we have tried for activation layers including ReLu, Sigmoid and tanh, but supurisingly the ReLu works the best. In this way, we still use ReLu for the activation function for all intermediate Dense layers.

\subsection{Different Agents Cooperate in Same Environments}

In this group of experiments, we put agent A and agent B into two identical environments respectively. The same environment means that all kinds of parameters of cart and pole in the two environments are the same. However, due to the randomness of the environment itself and the randomness of agent's choice of action, agent A and agent B actually interact in 2 independent environments, and their states can not be the same in most cases.

\begin{figure}
    \centering
    \centerline{\includegraphics[scale=0.08]{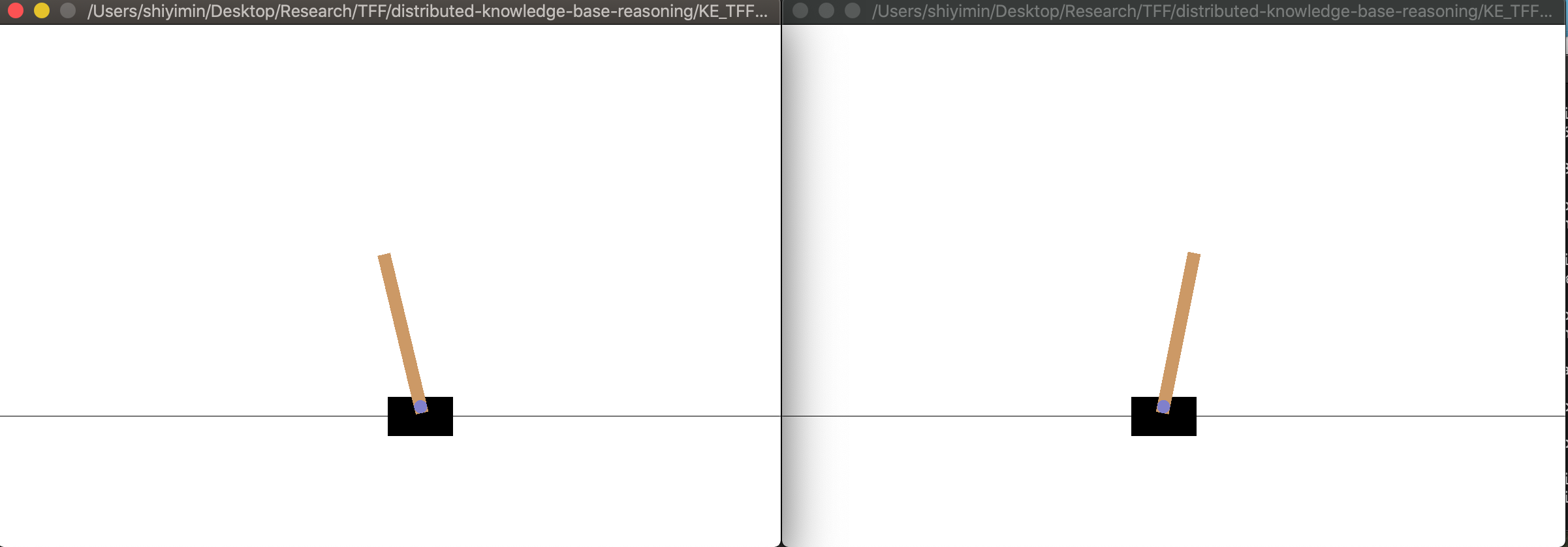}\includegraphics[scale=0.08]{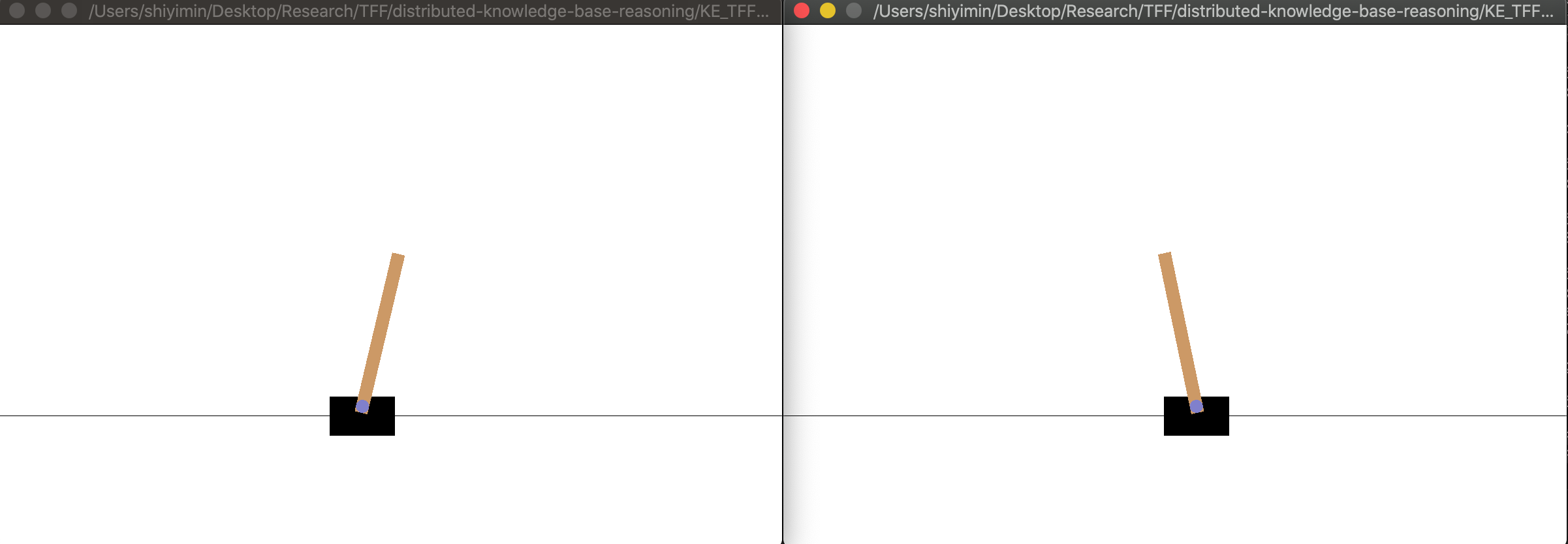}}
    \caption{Similar CartPole environments slightly differs in the coefficient of gravity.}
    \label{env:similar}
\end{figure}
\begin{figure}
    \centering
    \centerline{\includegraphics[scale=0.08]{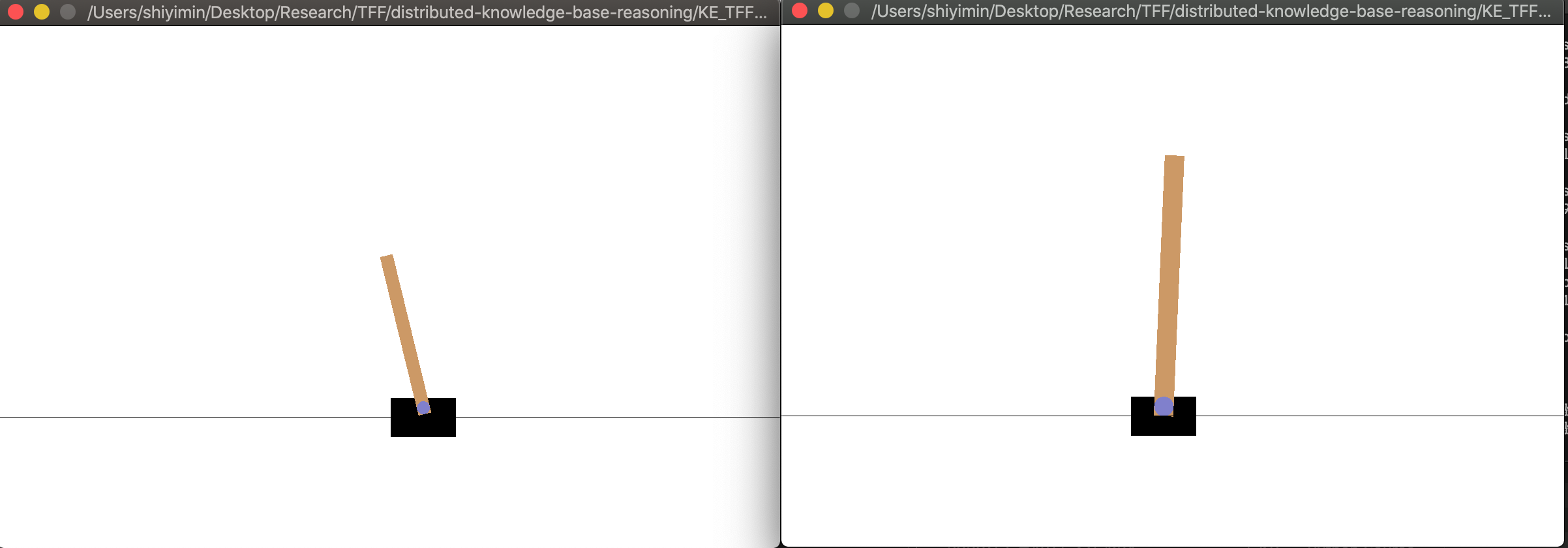}\includegraphics[scale=0.08]{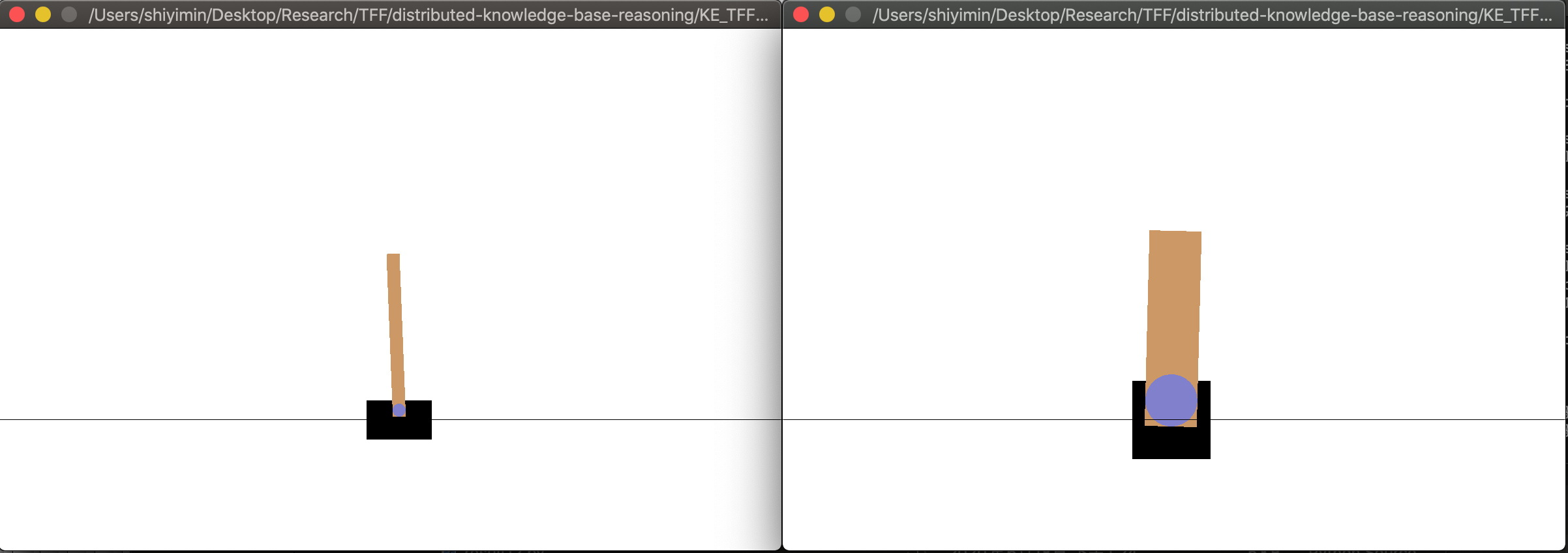}}
    \caption{CartPole environments differ in properties of the cart, pole and gravity.}
    % \label{env:similar}
\end{figure}

\begin{figure}
    \centering
    \centerline{\includegraphics[scale=0.5]{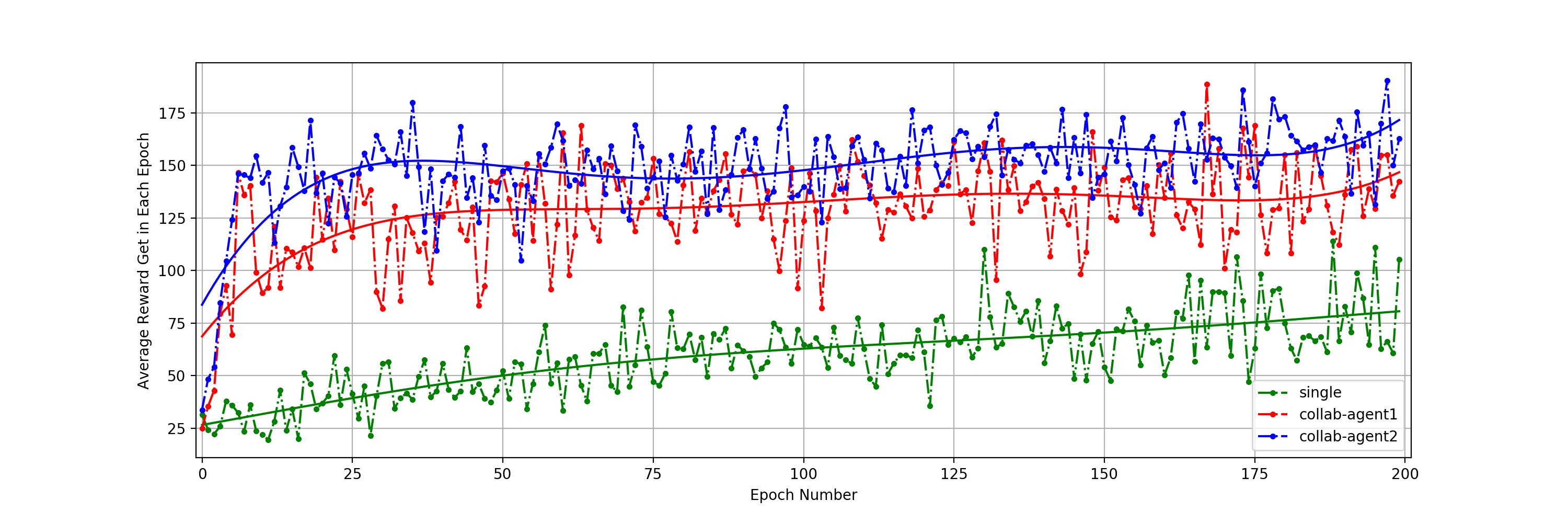}}
    \caption{Comparing "agents collaborate in same environments" with "agents train without collaboration".}
    \label{same}
\end{figure}

The red and blue curves shown in Fig.\ref{same} show the average return of the two agents in each epoch during 10 cooperative training processes, while the green curve shows the average return of each epoch during a single non-cooperative training process. We find that when two agents start to cooperate with each other according to our training framework, their average reward per round can be greatly improved to more than 120 in the first 30 epochs, while the agent trained alone can not even reach 100 in the $200^{th}$ epoch. We find that when two agents (in same environments) maintain and update the global layer by exchanging gradients environment, their fitting or converging speed will be much faster than the training process of a single agent without collaboration.

What's more, the first-hand agent A who firstly interact with its own environment and firstly update the shared layers could receive a relatively better performance and another second-hand agent B.

\subsection{Different Agents Cooperate in Similar but Slightly Different Environments}

\begin{figure}
    \centering
    \centerline{\includegraphics[scale=0.5]{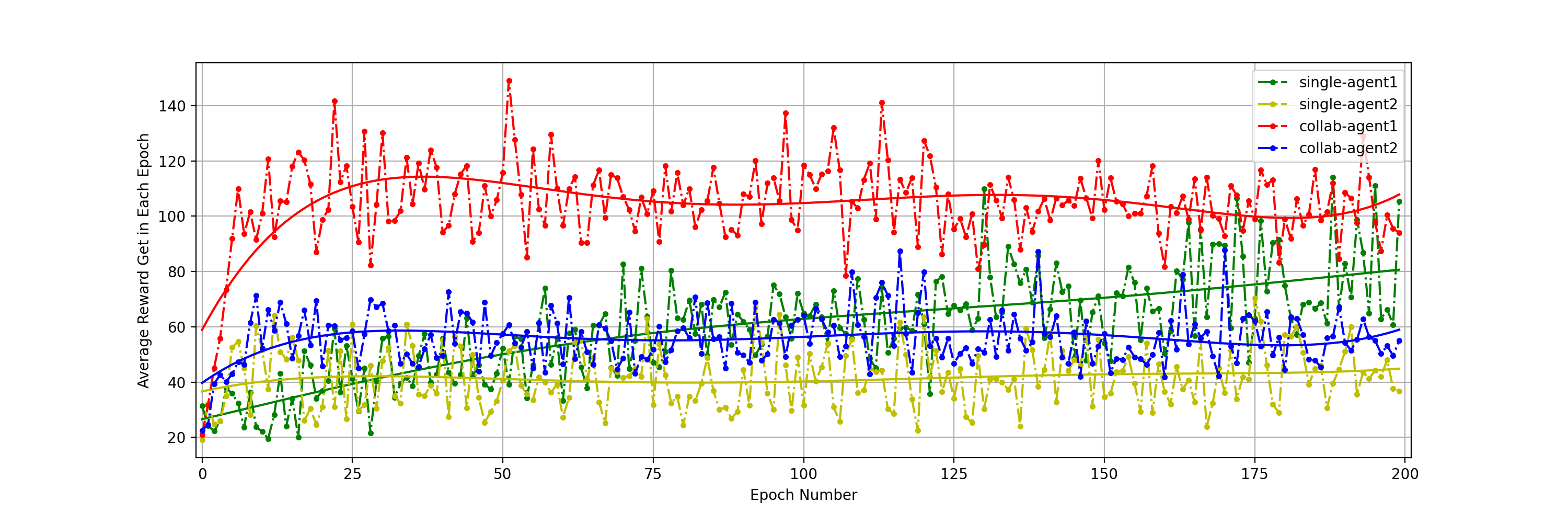}}
    \caption{Comparing "agents collaborate in similar but different environments" with "agents train without collaboration".}
    \label{similar}
\end{figure}

In this group of experiments, we put agent A and agent B into two similar but different environments respectively, which is obtained by modifying the parameters of Gym's environment configuration files and Python source files. In this group of experiments, we mainly modified the gravity parameters slightly, basically from 9.8 to 12.0. But we did not modify the properties of cart or pole, so we said that these are two similar but slightly different environments.

As before, the red and blue curves shown in the Fig.\ref{similar} are the average rewards of the two agents (A and B) in each epoch during 10 cooperative training processes, while the green curve is the average rewards of agent A of each epoch during a single non-cooperative training process, while the yellow curve stands for the condition that agent B complete the training process only by itself. We find that when two agents start to cooperate with each other according to our training framework, the average reward of agent A who first interacts with the environment and updates his local and global parameters in each epoch can be improved to more than 100 in the first 20 epochs, while the training result of agent B who updates later is not that ideal, even after about 60 epochs, it lags behind the reward condition of non-cooperative agent A, but it's still better than a single agent B without collaboration. We find that when two agents maintain and update the global layer through transmission gradient in similar but different environments, the order in which they interact with the environment and update the model parameters in each epoch will much more greatly affect their training results than in the sane-environment case. The firstly updated agent a will not perform as well as the cooperative agent in the same environment, but it is still far better than the agent training alone.

\subsection{Different Agents Cooperate in Obviously Different Environments}
\begin{figure}
    \centering
    \centerline{\includegraphics[scale=0.5]{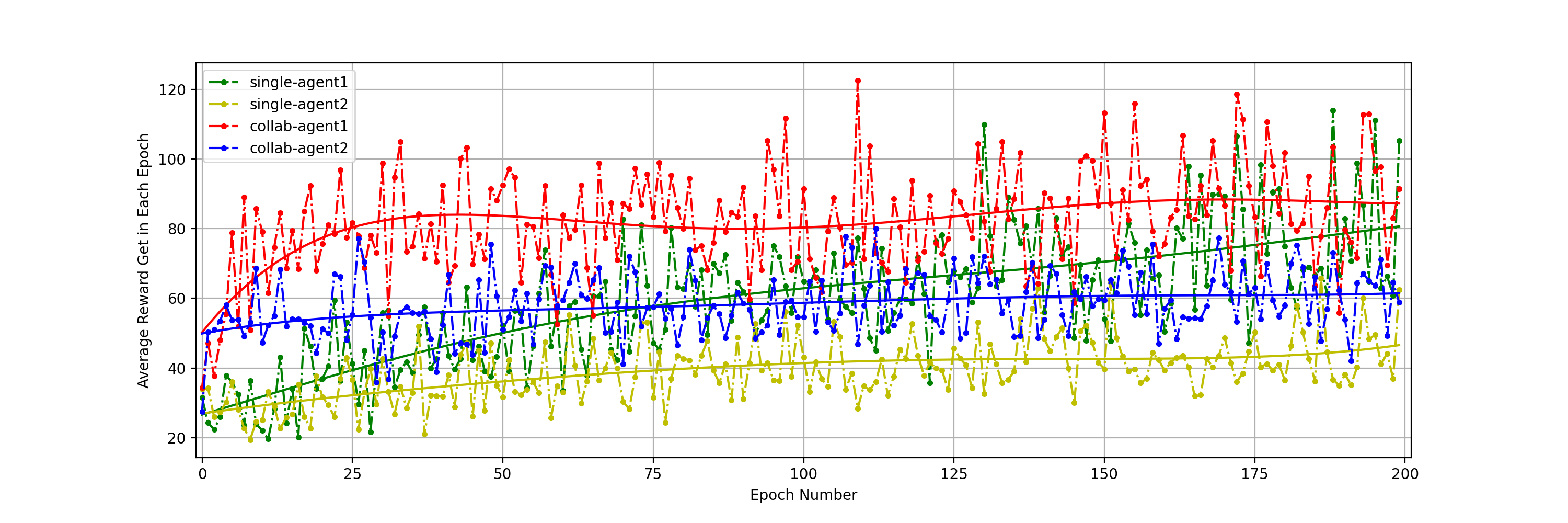}}
    \caption{Comparing "agents collaborate in more different environments (taller pole)" with "agents train without collaboration". The difference appears in properties of the cart, pole and gravity.}
    \label{tall}
\end{figure}
\begin{figure}
    \centering
    \centerline{\includegraphics[scale=0.5]{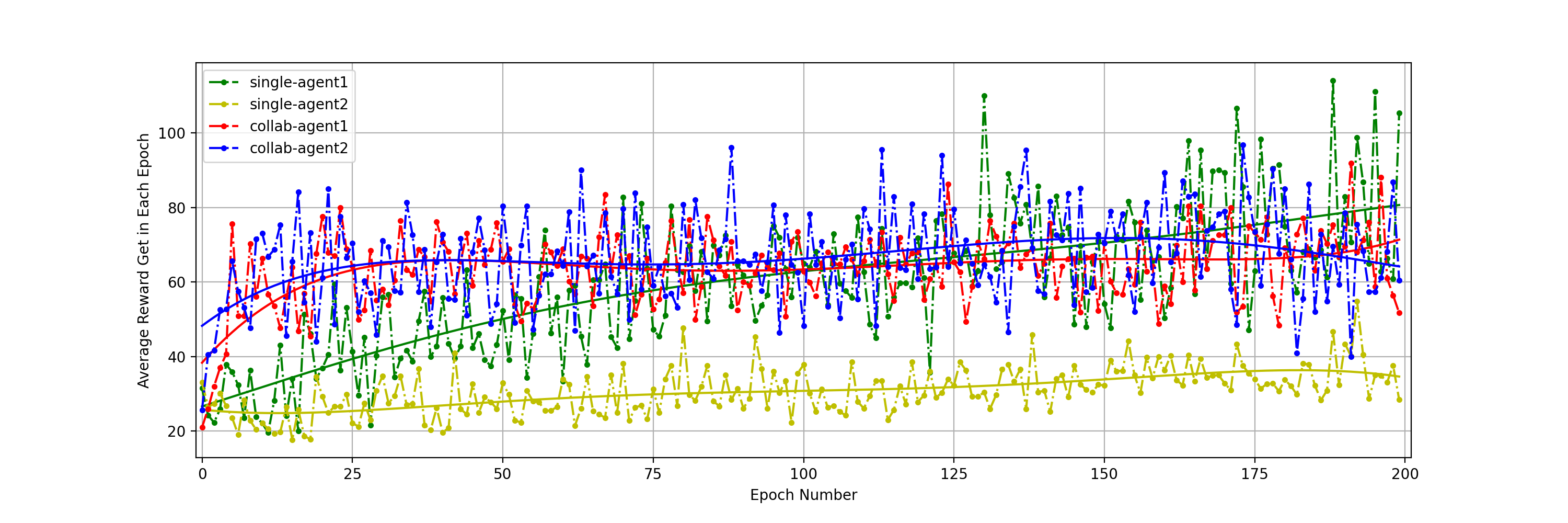}}
    \caption{Comparing "agents collaborate in another more different environments (fatter pole)" with "agents train without collaboration". The difference appears in properties of the cart, pole and gravity.}
    \label{fat}
\end{figure}
In this group of experiments, we put agent A and agent B into two groups of environments respectively. In each group of environments, agent A and agent B are in two more distinct environments with more obvious difference. In this group of experiments, we mainly modified the characteristics of cart and pole. In the first group of sub experiments, we mainly lengthen the length of pole while changing the gravity parameters. In the second group of sub experiments, we increase the gravity parameters, shorten the length of pole and increase the width of pole. So we say it's two distinctly different environments.

The results of the first group of experiments are shown in Fig.\ref{tall}. The red and blue curves shown in the figure are the average return of each epoch obtained by the two agents in 10 cooperative training processes, while the green and yellow curve is still the average return of each epoch obtained by these 2 agents if they are trained without collaboration respectively. We find that when two agents begin to cooperate with each other according to our training framework, and the environment is quite different, agent A who firstly interacts with the environment and updates its local and global parameters in each epoch will still be significantly increase more than agent B in its performance, and can stabilize above 80 after about 25 epochs, The training result of agent B, which is updated later, is not that ideal, but still better than the case where the agent B is trained alone. After about 60 epochs, agent B lags behind the non-cooperative agent A. However, the first updated agent A will not perform as well as the cooperative agents in same or similar environments, but it is still far better than the agent A training alone.

The results of the second group of experiments are shown in Fig.\ref{fat}. We find a similar observation that when the environment of two agents is quite different, although the training effect of agent A is similar to that of agent B, it can only be stable for more than 60 after about 25 epochs, which lags behind all the previous cooperative training agent A, but is better than the non-cooperative agent A.

\subsection{Different Agents Cooperate in Totally Different Environments}
\begin{figure}
    \centering
    \centerline{\includegraphics[scale=0.5]{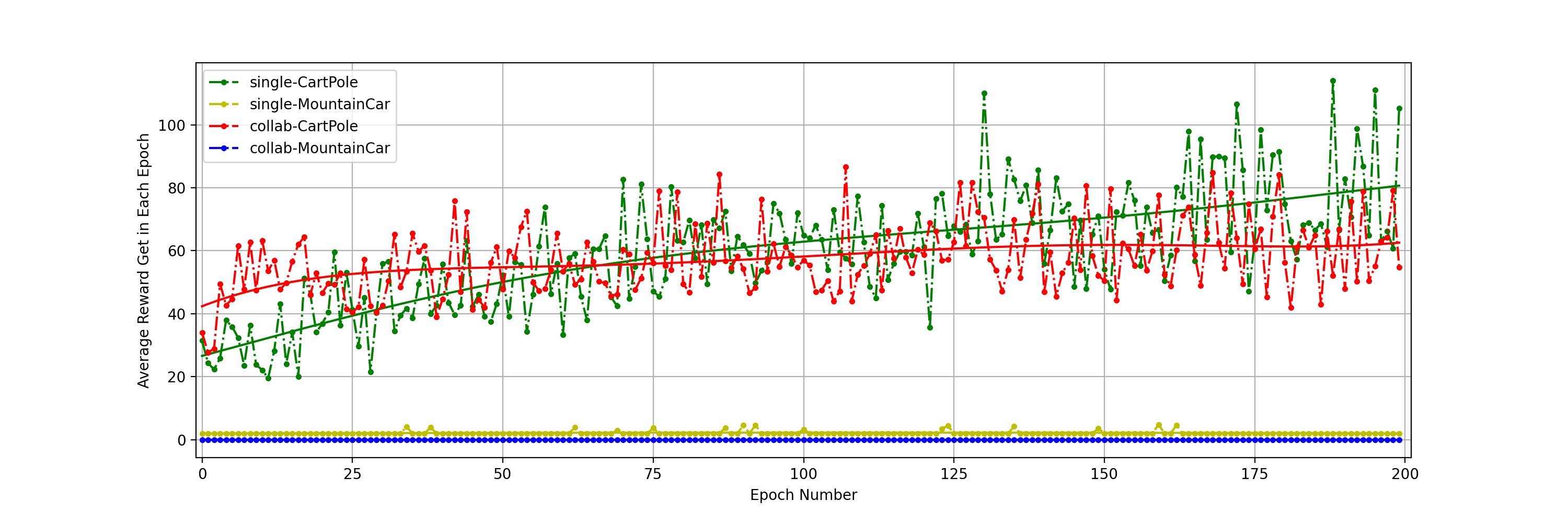}}
    \caption{Comparing "agents collaborate in totally different environments " with "agents train without collaboration". These two environments shares no common knowledge, one is "CartPole" and another one is "MountainCar".}
    \label{total}
\end{figure}

In this group of experiments, we put agent A and agent B into two completely different environments, in which the environment agent A locates is still CartPole, but the environment of agent B becomes a (artificially modified) MountainCar with the same observation space and action space dimensions. This means that two agents have no knowledge can share with each other. In this experiment, we want to see the worst result of the cooperation between the two agents.

The experimental results are shown in Fig.\ref{total}. The red and blue curves shown in the figure are the average return of the two agents in each epoch during 10 cooperative training sessions, while green and yellow curves are still the average return of each epoch during a single non-cooperative training process for agent A and B. We find that when two agents begin to cooperate with each other according to our training framework and the environment they are in has no common thing, agent B's training is completely invalid, and his reward is always 0 (in the classic version, it is - 200), which means that he has never been able to successfully climb the hill. Agent A can be stable at more than 50 after about 25 epochs, and can be trained by non-cooperative agent alone after about 60 epochs. This result is still slightly better than our expectation, because we see that agent A can still train non-cooperative agents better than single agent at least when we only allow training less than 60 epochs.

\begin{figure}
    \centering
    \centerline{\includegraphics[scale=0.5]{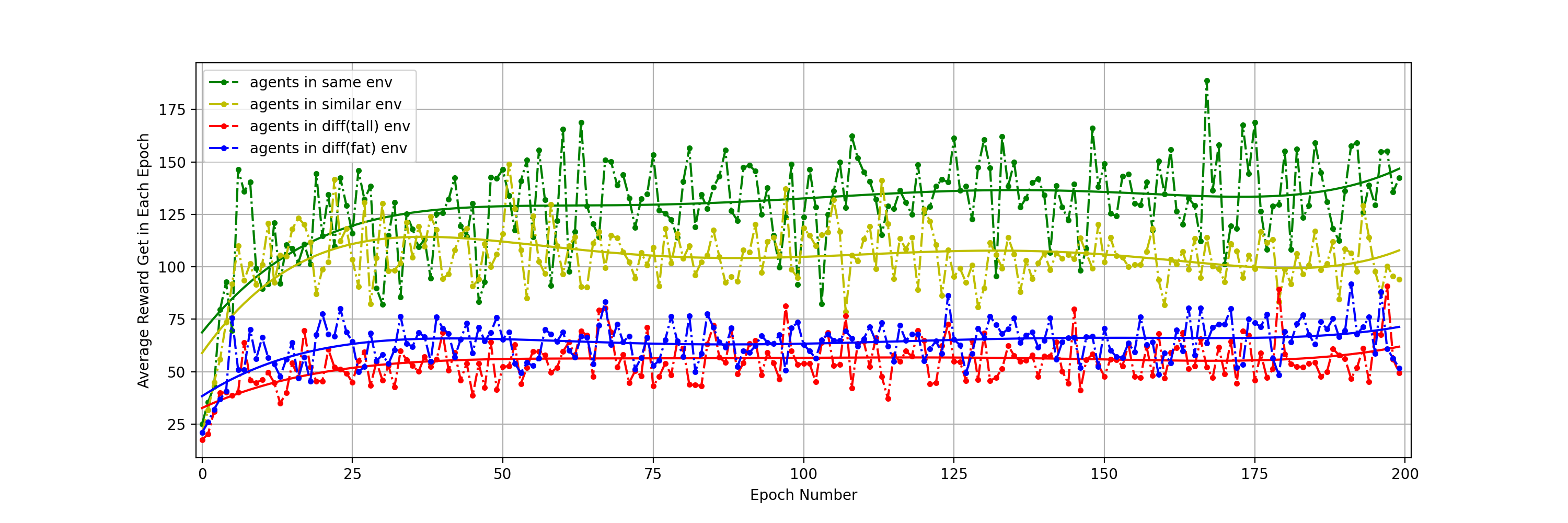}}
    \caption{Comparing the performance of Agent A when it cooperates with Agent B in different environment sets.}
    \label{compare}
\end{figure}

\subsection{Compare Cooperating Agents in Different Environments Groups}
Here by comparing the training curves of the different agent As in Fig.\ref{compare}, we can see that the training performance of the agent A with priority training is significantly decreased when the differences in environment become greater and greater. When agent A is cooperating with an agent B in a same environment, it will achieve a best performance, because they share a lot of common knowledge to learn from each other. When the agent B is in a similar but slightly different environment, the performance of agent A will be a little bit worse. If the environment where agent B locates is obviously different from A, but not totally different from it, the common knowledge can be more limited and cause the performance of A to be even worse. But can still be better than the situation where the agent A is trained by itself without any collaboration as long as the expected epoch number is not that large.

\section{Conclusions and Further Problems}
We can see that our cooperative network structure design scheme and distributed training framework can help the owners of different environments to achieve better convergence speed and performance in the process of training in many cases, and can also effectively ensure that different partners will not disclose privacy to each other. 

However, at present, there are still some problems worth further studying. First, the relationship between the training performance of agents and their training order. Second, the performance of agents when there are more-than-two participants in this system. For example, what will happen to each agent's performance if two of them are located in similar environments but the third lies in a totally different one? Third, what's the possible privacy attack based on this training framework? Whether the agents can attack each other in this non-central training system? And how to further prove the privacy protection of it? 

\bibliographystyle{unsrt}  
\bibliography{references}  %%% Remove comment to use the external .bib file (using bibtex).


% George Kour and Raid Saabne.
% \newblock Real-time segmentation of on-line handwritten arabic script.
% \newblock In {\em Frontiers in Handwriting Recognition (ICFHR), 2014 14th
%   International Conference on}, pages 417--422. IEEE, 2014.

% George Kour and Raid Saabne.
% \newblock Fast classification of handwritten on-line arabic characters.
% \newblock In {\em Soft Computing and Pattern Recognition (SoCPaR), 2014 6th
%   International Conference of}, pages 312--318. IEEE, 2014.

% \bibitem{hadash2018estimate}
% Guy Hadash, Einat Kermany, Boaz Carmeli, Ofer Lavi, George Kour, and Alon
% \newblock Estimate and replace: A novel approach to integrating deep neural
%   networks with existing applications.
% \newblock {\em arXiv preprint arXiv:1804.09028}, 2018.

% \end{thebibliography}

\end{document}